\documentclass[runningheads]{llncs}

\usepackage{eccv}

\usepackage{eccvabbrv}

\usepackage{graphicx}
\usepackage{booktabs}

\usepackage[accsupp]{axessibility}  

\usepackage{hyperref}

\usepackage{orcidlink}
\usepackage{eccvabbrv}
\usepackage{indentfirst}
\usepackage{lineno}
\usepackage{subcaption}
\usepackage{arydshln}
\usepackage{multirow}
\usepackage{longtable}
\usepackage{color}
\usepackage{colortbl}
\usepackage{cite}
\usepackage{booktabs}
\usepackage{graphicx}
\usepackage{stfloats}
\usepackage{amssymb,amsmath}
\usepackage{epstopdf}
\usepackage{rotating}
\usepackage{threeparttable}
\usepackage{bbding}
\usepackage{comment}

\usepackage{algorithmicx}%
\usepackage{algpseudocode}%
\usepackage{amsmath,amsfonts}
\usepackage{algorithm}
\usepackage{multirow}

\newcommand{\M}[1]{\mathbf{#1}}

\begin{document}

\title{Diff-Reg: Diffusion Model in Doubly Stochastic Matrix Space for Registration Problem} 

\titlerunning{Diff-Reg}

\author{Qianliang Wu\inst{1}\orcidlink{0000-0001-6592-021X}\and
Haobo Jiang\inst{5} \and
Lei Luo\inst{1} \and Jun Li\inst{1} \and Yaqing Ding\inst{4}\orcidlink{0000-0002-7448-6686} \and \\Jin Xie\inst{2,3}\textsuperscript{\Envelope} \and Jian Yang\inst{1}\orcidlink{0000-0003-4800-832X}\textsuperscript{\Envelope}}

\authorrunning{Wu et al.}

\institute{PCA Lab, Key Lab of Intelligent Perception and Systems for High-Dimensional Information of Ministry of Education, and Jiangsu Key Lab of Image and Video Understanding for Social Security, School of Computer Science and Engineering, Nanjing University of Science and Technology, Nanjing, China \and
State Key Laboratory for Novel Software Technology, Nanjing University, Nanjing, China \and
School of Intelligence Science and Technology, Nanjing University, Suzhou, China \and
Visual Recognition Group, Faculty of Electrical Engineering, 
	Czech Technical University in Prague, Prague, Czech Republic \and National University of Singapore, Singapore\\
    \email{{wuqianliang}@njust.edu.cn}}

\maketitle
\makeatletter\def\Hy@Warning#1{}\makeatother
\let\thefootnote\relax\footnotetext{\textsuperscript{\Envelope} Corresponding Authors}

\begin{abstract}
Establishing reliable correspondences is essential for 3D and 2D-3D registration tasks. Existing methods commonly leverage geometric or semantic point features to generate potential correspondences. However, these features may face challenges such as large deformation, scale inconsistency, and ambiguous matching problems (e.g., symmetry). Additionally, many previous methods, which rely on single-pass prediction, may struggle with local minima in complex scenarios. To mitigate these challenges, we introduce a diffusion matching model for robust correspondence construction. Our approach treats correspondence estimation as a denoising diffusion process within the doubly stochastic matrix space, which gradually denoises (refines) a doubly stochastic matching matrix to the ground-truth one for high-quality correspondence estimation. It involves a forward diffusion process that gradually introduces Gaussian noise into the ground truth matching matrix and a reverse denoising process that iteratively refines the noisy one. In particular, we deploy a lightweight denoising strategy during the inference phase. Specifically, once points/image features are extracted and fixed, we utilize them to conduct multiple-pass denoising predictions in the reverse sampling process. Evaluation of our method on both 3D and 2D-3D registration tasks confirms its effectiveness. The code is available at \href{red}{https://github.com/wuqianliang/Diff-Reg}.
\end{abstract}

\section{Introduction}
The 3D registration problem, encompassing point cloud registration and image-to-point cloud registration, is critical in various computer vision and computer graphics applications, including 3D reconstructions, localization, AR, and robotics. These applications usually necessitate precise correspondences (matchings) between point cloud pairs or image-to-point cloud pairs for reliable rigid transformation or non-rigid deformation estimations.

The goal of achieving accurate matchings is to identify the most significant correspondences~\cite{zhong2009intrinsic,yu2023rotation, bai2020d3feat} with local or global semantic or geometric consistency~\cite{li20232d3d,deng2018ppfnet,qin2022geometric,wu2023sgfeat}. However, this objective would be challenging, especially in situations with globally ambiguous matching patches, large deformation, scale inconsistency, and low overlapping problems.

Recently, deep learning-based feature matching methods~\cite{qin2022geometric,li2022lepard,yew2022regtr, mei2023unsupervised,huang2021predator,yu2021cofinet,yu2023rotation,yao2023hunter,li20232d3d} have achieved significant progress in point cloud registration by employing UNet-like~\cite{thomas2019kpconv,he2016deep} backbones to extract superpoints (subsampled patches) and their associated features. These methods typically compute an initial matching matrix between superpoints in the feature space. Additionally, outlier rejection techniques~\cite{yang2020teaser, bai2021pointdsc,chen2022sc2,jiang2023robust,zhang20233dmac} propose specialized methodologies to identify improved inlier correspondences based on certain semantic or geometric priors~\cite{zhong2009intrinsic,wu2023sgfeat,qi2019deep,fu2021robust,yan2024tri,yan2023desnet,yan2022learning,yan2022rignet}. However, these methods usually rely on a single-pass prediction of correspondences, which may not always yield optimal results.

In this paper, drawing inspiration from the diffusion model~\cite{ho2020denoising, song2020denoising, vignac2022digress,austin2021structured}, we introduce a diffusion matching model in the doubly stochastic matrix space\cite{86bd1ad6-50bb-38d0-9978-0966b4dfc6d3}. By training a diffusion model with the doubly stochastic matrix space as a feasible solution domain, we effectively learn a generalized optimization algorithm specifically designed for the doubly stochastic matrix space, adapted to the characteristics of the dataset or scene. Our diffusion matching model consists of two main components: a forward diffusion process and a reverse denoising process, which operates within the matrix space. The forward diffusion process gradually introduces Gaussian noise into the ground truth matching matrix, while the reverse denoising process iteratively refines the noisy matrix to the optimal one. 
For efficiency, we propose a novel and generalized lightweight denoising module that can be adapted to 2D-3D and 3D registration tasks. Finally, we establish a specific variational lower bound associated with our diffusion matching model in the doubly stochastic matrix space and a simplified version of the objective function to train our framework effectively.

\textbf{Why diffused in the doubly stochastic matrix space?}
Tasks like image or point cloud registration face challenges like scale inconsistency, large deformation, ambiguous matching, and low overlapping. Several state-of-the-art studies~\cite{qin2022geometric,yao2023hunter,wu2023sgfeat,wugraph,mei2021cotreg,bai2021pointdsc,zhang20233dmac} have attempted to encode high-order combinational geometric consistency. However, these manually crafted designs may not encompass all potential effective strategies for various challenging scenarios (e.g., large deformation). The diffusion process in the matrix space is a practical data augmentation technique that can generate additional training samples incorporating any-order combinational geometric consistency, offering a promising approach to address these challenges. A doubly stochastic matching matrix is, in fact, a dual-directional mapping. It offers a one-to-one mapping relation constraint for any kind of two-view (and any two-modality) matching/registration problem. The diffusion process within the matching matrix space naturally provides a broader range of training samples. Furthermore, the many-to-one relationship between the matching matrix space and the warping operation space can facilitate our method in learning a more optimal sampling path used in the reverse sampling process, in contrast to methods that conduct diffusion in SE(3) space.

Our approach presents several advantages compared to previous registration methods. The forward diffusion process generates diverse training samples, acting as data augmentation for the feature backbone and single-pass prediction head. The drawback of single-pass prediction methods is that if correspondences are predicted in a local minimum, subsequent outlier rejection or post-processing steps may face significant challenges. In contrast, our reverse denoising process, guided by the posterior distribution, allows for escaping from local minima, enabling the process to initiate from either white noise or any initial solution. Another enhancement is eliminating the feature backbone during the reverse denoising process at inference time. This streamlined design enables the denoising sampling process to explore a broader solution space (e.g., the matrix space), increasing diversity and facilitating more iterative steps. The findings from our empirical experiments support these claims. 

Our contributions are summarized as follows:
\begin{itemize}
    \item To our knowledge, we are the first to deploy the diffusion model in the doubly stochastic matrix space for iteratively exploring the optimal matching matrix through the reverse denoising sampling process.
    
    \item The lightweight design of our reverse denoising module results in faster convergence in the reverse sampling process. Moreover, our framework can effectively utilize reverse denoising sampling in a noise-to-target fashion or start from a highly reliable initial solution.

    \item We conducted comprehensive experiments on the real-world 4DMatch~\cite{li2022non}, 3DMatch~\cite{zeng20173dmatch}, and RGB-D Scenes V2~\cite{zeng20173dmatch,li20232d3d} datasets to validate the effectiveness of our diffusion matching model on 3D registration and 2D-3D registration task.
\end{itemize}

\section{Related work}

\subsection{3D and 2D-3D Registration}

The registration problem estimates the transformation between the point cloud or image-to-point-cloud pair. Recently, there have been significant advancements in feature learning-based methods for point cloud registration. Many of these state-of-the-art approaches, such as~\cite{bai2020d3feat,huang2021predator,yu2021cofinet,qin2022geometric,yu2022riga,yu2023peal}, leverage a backbone architecture similar to KPConv~\cite{thomas2019kpconv} to downsample points and generate features with larger receptive fields. To further enhance the performance of these methods, they integrate prior knowledge and incorporate learnable outlier rejection modules. For instance, GeoTR~\cite{qin2022geometric} introduces angle-wise and edge-wise embeddings into the transformer encoder, while RoITr~\cite{yu2023rotation} integrates local Point Pair Features (PPF)~\cite{deng2018ppf} to improve rotation invariance.

In addition to feature learning-based methods, another category of registration methods focuses on outlier rejection of candidate correspondences. For instance, PointDSC~\cite{bai2021pointdsc} utilizes a maximum clique algorithm in the local patch to cluster inlier correspondences. SC2-PCR~\cite{chen2022sc2} constructs a second-order consistency graph for candidate correspondences and theoretically demonstrates its robustness. Building on the second-order consistency graph proposed by SC2-PCR~\cite{chen2022sc2}, MAC~\cite{zhang20233dmac} introduces a variant of maximum clique algorithms to generate more reliable candidate inlier correspondences. Moreover, methods such as PEAL~\cite{yu2023peal} and DiffusionPCR~\cite{chen2023diffusionpcr} employ an iterative refinement strategy to enhance the overlap prior information obtained from a pre-trained GeoTr~\cite{qin2022geometric}. 

Recently, significant advancements have been made in 2D-3D registration methods~\cite{Li2021DeepI2PIC, Wang2021P2NetJD, Wang2023FreeRegIC,li20232d3d}. These methods face similar challenges to 3D registration tasks, with the additional complexity of scale inconsistency caused by the perspective projection of images. To address the issue of scale inconsistency, we propose the incorporation of a pre-trained feature backbone, DINO v2~\cite{oquab2023dinov2}, which offers superior multiscale features. Additionally, implementing diffused data augmentation in our diffusion matching model can enhance the ability to identify prominent combinational and consistent correspondences.

\subsection{Diffusion Models for 3D Registration}
Recently, the diffusion model~\cite{ho2020denoising,song2019generative,song2020denoising} has made great development in many fields, including human pose estimation~\cite{Gong2022DiffPoseTM, Shan2023DiffusionBased3H}, camera pose estimation~\cite{Wang2023PoseDiffusionSP}, object detection~\cite{chen2023diffusiondet}, segmentation~\cite{Baranchuk2021LabelEfficientSS, Gu2022DiffusionInstDM}. These developments have been achieved through a generative Markov Chain process based on the Langevin MCMC~\cite{parisi1981correlation} or a reversed diffusion process~\cite{song2020denoising}. Recognizing the power of the diffusion model to iteratively approximate target data distributions from white noise using hierarchical variational decoders, researchers have started applying it to point cloud registration and 6D pose estimation problems. 

The pioneer work~\cite{urain2023se} that applied the diffusion model in the SE(3) space was accomplished by utilizing NCSN~\cite{song2019generative} to learn a denoising score matching function. This function was then used for reverse sampling with Langevin MCMC in SE(3) space to evaluate 6DoF grasp pose generation. Additionally, \cite{jiang2023se} implemented DDPM~\cite{ho2020denoising} in the SE(3) space for 6D pose estimation by employing a surrogate point cloud registration baseline model. Similarly, GeoTR~\cite{qin2022geometric} served as a denoising module in~\cite{chen2023diffusionpcr}, gradually denoising the overlap prior given by the pre-trained model, following a similar approach to PEAL~\cite{yu2023peal}.

\section{The proposed Approach}

\subsection{Problem Formulation}
Given source point clouds $\M P \in \mathbb{R}^{N{\times}3}$ and target point clouds $\M Q \in \mathbb{R}^{M{\times}3}$, the 3D registration task is to find top-k correspondences $\mathcal{C}$ from matching matrix $\M E$ and to conduct warping transformation ($\M{\Gamma} \in$ SE(3) for rigid transformations, and 3D flow fields for non-rigid transformations) to align the overlap region of $\M P$ and $\M Q$. In the context of 2D-3D registration, with a source image $\M X \in \mathbb{R}^{H{\times}W{\times}2}$ and target point cloud $\M Y \in \mathbb{R}^{M{\times}3}$, the standard pipeline involves determining the top-k correspondences $\mathcal{C} = \{(\M x_i,\M y_j)|\M x_i \in \mathbb{R}^2, \M y_j \in \mathbb{R}^3\}$, and then estimating the rigid transformation $\M{\Gamma} \in$ SE(3) by minimizing the 2D projection error:
\begin{equation}
\nonumber
    \underset{\M{\Gamma} \in SE(3)}{\min} \sum \limits_{\M x_i,\M y_j\in \mathcal{C}} ||Proj(\M{\Gamma}(\M y_j),\M K)-\M x_i||_2
\end{equation}
where $\M K$ represents the camera intrinsic matrix, and $Proj(\cdot, \cdot)$ denotes the projection function from 3D space to the image plane.

\begin{figure*}
    
      \centering
      \includegraphics[width=11cm, height=7cm]{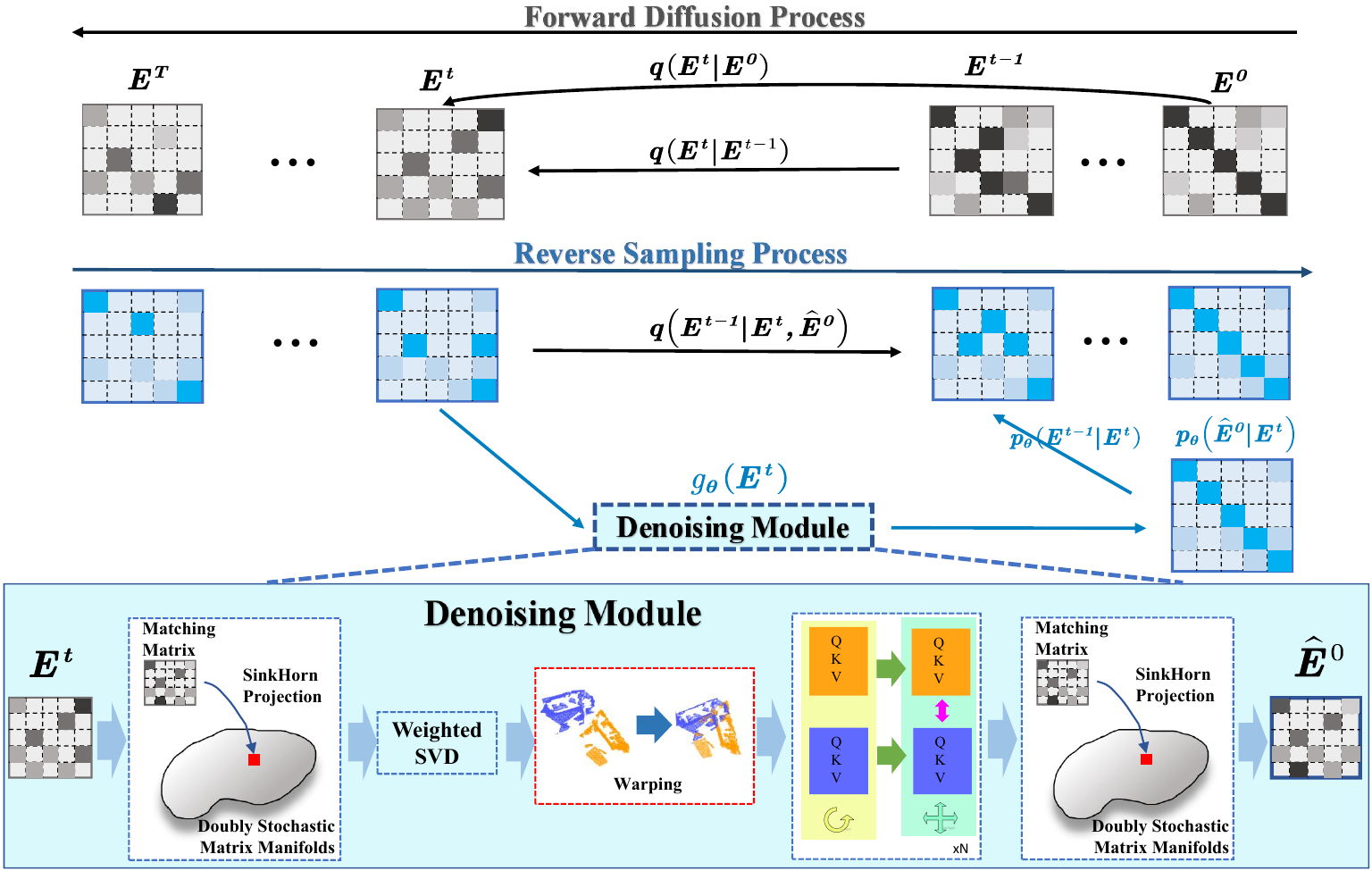}
    
\caption{Overview of our diffusion matching model. The forward diffusion process is driven by the Gaussian transition kernel $q(\M E^t|\M E^{t-1})$, which has a closed form $q(\M E^t|\M E^0)$. The denoising model $g_\theta(\M E^t)$ learns a reverse denoising gradient that points to the target solution $\M E^0$. During inference, in the reverse sampling process, we utilize the predicted $\hat{\M E}_0$ and DDIM~\cite{song2020denoising} to sampling $\M E^{t-1}$.} \label{framework}    

\end{figure*}

\subsection{Overview}
Our framework comprises a feature backbone (e.g., KPConv~\cite{thomas2019kpconv}/ResNet~\cite{he2016deep}) and a diffusion matching model~\cite{ho2020denoising}. In the 3D registration task, the KPConv backbone takes source point clouds $\M P$ and target $\M Q$ as input and performs downsamplings to obtain the superpoints $\hat{\M P}$ and $\hat{\M Q}$, along with their associated features $\M F^{\hat{\M P}} \in \mathbb{R}^{N{\times}d}$ and $\M F^{\hat{\M Q}} \in \mathbb{R}^{M{\times}d}$. In the 2D-3D registration task, a ResNet~\cite{he2016deep} with FPN~\cite{Lin2016FeaturePN} downsamples the image $\M X\in \mathbb{R}^{H{\times}W{\times}2}$ (in image-point-coud pair $(\M X, \M Q)$) to the superpixels $\hat{\M X}\in \mathbb{R}^{\hat{H}{\times}\hat{W}{\times}2}$ and give the associated image feature $\M F^{\hat{\M X}} \in \mathbb{R}^{\hat{H}{\times}\hat{W}{\times}d}$, while the target point cloud $\M Q$ is processed by a KPConv backbone, similar to the 3D registration task. Additionally, a depth map $\M D^{\hat{\M X}}$ and new superpixel feature $\M F^{\hat{\M X}}_{dino}$ for superpixels $\hat{\M X}$ are given by the pre-trained depth estimation model~\cite{yang2024depth} and visual feature backbone DINO v2~\cite{oquab2023dinov2}, respectively. 

Our diffusion module $g_\theta$ (refer to Section \ref{dmd}) takes these superpoints (or superpixels) and associated features as inputs. We employ one loss for our denoising module and another for the single-pass prediction head during the training stage. During inference, we utilize $g_\theta$ in the reverse sampling process to predict the target matching matrix $\M E^0$ from any noisy one and exploit DDIM~\cite{song2020denoising} to sample a more accurate matching matrix. The denoising module $g_{\theta}$ primarily consists of four components: (1) Sinkhorn Projection~\cite{cuturi2013sinkhorn}, (2) Weighted SVD~\cite{besl1992method}, (3) Warping Function (4) Denoising Transformer Network~\cite{vaswani2017attention} and (5) Matching function. More details of our framework can be found in section \ref{dmd} and the appendix.

\subsection{Diffusion Model in Doubly Stochastic Matrix Space}\label{continuous_diffusion}
In this section, we introduce the construction of our diffusion matching model for generating the matching matrix between two scans. We denote the matching matrix as $\M E\in \{0,1\}^{N{\times}M}$, and we assume $\M E$ is defined in a nonsquare ``doubly stochastic'' matrix space $\mathcal{M}$ (refer to Appendix).

\noindent\textbf{Forward Diffusion Process.}
As mentioned in DDPM~\cite{ho2020denoising}, the forward diffusion process is fixed to a Markovian chain, denoted as $q(\M E^{1:T}|\M E^0)$, which generates a sequence of latent variables $\M E^t$ by gradually injecting Gaussian noise into the ground truth matching matrix $\M E^0$. The diffused matching matrix $\M E^t$ at arbitrary timestamp \textit{t} has a closed form:
\begin{eqnarray}
    \M E^t \sim q(\M E^t|\M E^0) = \mathcal{N}(\M E^t;\sqrt{\bar{\alpha}}\M E^0,(1-\bar{\alpha})\textbf{I}))\label{diff_10}.
\end{eqnarray}
where the added noise over each element of the matrix is sampled independently and identically distributed (i.i.d.). 
However, this diffused $\M E^t \sim q(\M E^t|\M E^0)$ is a continuous matrix in $\mathbb{R}^{N \times M}$, which is outside the feasible solution space of matching matrices (i.e., doubly stochastic matrix manifolds). To address this issue, we apply the following projection to confine the matrix $\M E^t$ to the feasible solution space $\mathcal{M}$:
\begin{eqnarray}
\begin{aligned}
\rm{(Rigid)}\quad &\M E^t = \sqrt{\bar{\alpha_t}}\M E^0 + \sqrt{1-\bar{\alpha_t}}f_{\epsilon}(\epsilon_0), \quad \tilde{\M E}^t = \M E^t - \rm{Min}(\M E^t), \\
\rm{(Deformable)}\quad &\M E^t = \sqrt{\bar{\alpha_t}}\M E^0 + \sqrt{1-\bar{\alpha_t}}\epsilon_0, \quad \tilde{\M E}^t = \rm{Sigmoid}(\M E^t),\\
&\tilde{\M E}^t = \mathbf{f}_{\rm{sinkhorn}}(\tilde{\M E}^t) 
\end{aligned}
\end{eqnarray}
where the $\mathbf{f}_{\text{sinkhorn}}$ operation is from the Sinkhorn algorithm~\cite{cuturi2013sinkhorn} and $f_{\epsilon}=(\epsilon\%1)(abs(\epsilon)/\epsilon)\eta$. We empirically set $\eta=1.5$ and $\epsilon_0 \sim \mathcal{N}(\epsilon;0,\textbf{I})$.

\noindent\textbf{Reverse Denoising Sampling Process.}
Given a diffusion Markovian chain $\M E^0 \rightarrow \M E^1 \rightarrow ... \rightarrow \M E^T$, we need to learn a reverse transition kernel with the posterior distribution $q(\M E^{t-1}|\M E^t, \M E^0)$ to sample the reverse Markovian chain $\M E^T \rightarrow \M E^{T-1} \rightarrow ... \rightarrow \M E^0$ from a white noise $\M E^T$ to achieve the target matching matrix. The posterior distribution $q(\M E^{t-1}|\M E^t,\M E^0)$ conditioned on $\M E^0$ and $\M E^t$ is defined as: 
\begin{equation}
\begin{split}
    &q(\M E^{t-1}|\M E^t,\M E^0) = \frac{q(\M E^t|\M E^{t-1},\M E^0)q(\M E^{t-1}|\M E^0)}{q(\M E^t|\M E^0)}\\
    &\propto \mathcal{N}(\M E^{t-1};\underbrace{\frac{\sqrt{\alpha_t}(1-\bar{\alpha}_{t-1})\M E^t+\sqrt{\bar{\alpha}_{t-1}}(1-\alpha_{t})\M E^0}{1-\Bar{\alpha}_t}}_{\mu_q(\M E^t,\M E^0)},\underbrace{\frac{(1-\alpha_t)(1-\bar{\alpha}_{t-1})}{1-\bar{\alpha}_t}\textbf{I}}_{\Sigma_q(t)} ).
\end{split}\label{eqn_rds_matrix}
\end{equation} 

To effectively train our denoising network, we derive the variational lower bound of the log-likelihood of the training samples $\M E^0$:
\begin{eqnarray}
\begin{aligned}
    log p(\M E^0) \geq \mathbb{E}_{\M E^{1:T} \sim q(\M E^{1:T}|\M E^0)}\left[ log\left( \frac{p_{\theta}(\M E^{0:T})}{q(\M E^{1:T}|\M E^0)} \right) \right]
    \propto \mathbb{E}_q \left[\sum \limits_{t=2}^T log(p_\theta(\M E^0|\M E^t))\right].
\end{aligned}\label{elbo_matrix}
\end{eqnarray}
Based on the derivation of Eqn. (\ref{elbo_matrix}), we can further simplify the variational lower bound above to train $p_\theta(\M E^0|\M E^t)$:
\begin{eqnarray}
    L_{simple} = -\mathbb{E}_{q(\M E^0)}\left[\sum \limits_{t-1}^T\mathbb{E}_{q(\M E^t|\M E^0)}logp_\theta(\M E^0|\M E^t)\right]\label{L_simple}.
\end{eqnarray}

\subsection{The Lightweight Denoising Module $g_\theta$}\label{dmd}
This section outlines the architecture of the lightweight denoising module $g_\theta$. In 3D registration tasks, $g_\theta$ take the superpoints $\hat{\M P}$, $\hat{\M Q}$ with associated points features $\M F^{\hat{\M P}}$, $\M F^{\hat{\M Q}}$ as the inputs in the reverse sampling process. Similarly, in 2D-3D registration tasks, the inputs of $g_\theta$ are superpixels $\hat{\M X}$ and superpoints $\hat{\M Q}$ with associated features $\{\M F^{\hat{\M X}}, \M D^{\hat{\M X}}, \M F^{\hat{\M X}}_{dino}, \M F^{\hat{\M Q}}\}$. At inference time, $g_\theta$ inputs a noised matching matrix $\M E^t$ and outputs a predicted target matching matrix $\hat{\M E}_{0}$.

We define the denoising module $g_\theta$ by sequentially stacking five components as a differentiable layer:

\noindent\textbf{Sinkhorn Projection}: $\mathbf{f}_{\text{sinkhorn}}(\cdot)$\label{sinkhorn_projection}.
To constrain the matching matrix $\M E^t$ within the doubly stochastic matrices manifolds, we utilize the SinkHorn~\cite{cuturi2013sinkhorn} iterations to project $\M E^t$. We treat this operation as a key role in our framework rather than a post-processing in other methods.

\noindent\textbf{Weighted SVD}: $\mathbf{soft\_ procrustes}(\cdot,\cdot,\cdot)$\label{procrustes}.
Given top-k confident correspondences $\kappa$, we utilize the weighted SVD algorithm~\cite{arun1987least} (differentiable)
 to compute the transformation $\M R,\M t$ in a closed form:
 \begin{equation} 
 \begin{split}
     \M H&=\sum \limits_{(i,j)\in C} \Tilde{\M E}(i,j)\hat{\M p}_i\hat{\M q}_j^\top,\ \M H=\M U{\M \Lambda}\M V^\top,\\
     \mathbf R&=\M U {\rm diag}(1,1,\det(\M U \M V^\top))\M V,\\
     \mathbf t&=\frac{1}{|\kappa|}\left( {\sum \limits_{(i,\cdot)\in \kappa}\ \hat{\M p}_i} - \mathbf{R}{\sum \limits_{(\cdot,j)\in \kappa}}\hat{\M q}_j\right)
\end{split}
 \end{equation}
where $(\hat{\M p}_i,\hat{\M q}_j)$ is a superpoint correspondences. In the 2D-3D registration task, we first project a superpixel depth map $\M D^{\hat{\M X}}$ to a point cloud $\M P^{\hat{\M X}}_D$ using camera intrinsic. Then we utilize this SVD decomposition to compute $\mathbf{R},\mathbf{t}$ with $\M P^{\hat{\M X}}_D$ and $\hat{\M Q}$ as inputs. 

The Weighted SVD operation can be viewed as a differentiable projection that arises from the matching matrix $\Tilde{\M E}$ obtained from the first Sinkhorn Projection. Alternatively, in the 2D-3D registration task, this Weighted SVD operation can be substituted with a differentiable PnP (Perspective-n-Point)~\cite{li2012robust}.

\noindent\textbf{Warping Function}: $\mathbf{warping}(\cdot,\cdot,\cdot)$.
After obtaining transformation $\mathbf{R},\mathbf{t}$, the rigid warping of source point clouds is computed by $\M W(\hat{\M p}_i) = \mathbf{R}\hat{\M p}_i+\mathbf{t}$. In this paper, we use rigid warping for rigid and deformable registration cases to demonstrate our design. As for deformable registration, with the denoised correspondences, we can compute the flow fields for all points in $\hat{\M P}$ by performing nearest neighbor interpolation with the predicted inlier correspondences as anchors. 

\noindent\textbf{{Denoising Transformer}: $f_\theta(\cdot,\cdot,\cdot,\cdot,\cdot,\cdot)$}\label{denoising_transformer}. 
We observed empirically that a simple noise model does not hurt performance. Thus, we exploit a lightweight Transformer~\cite{vaswani2017attention} as our denoising network. Specifically, we utilize a 6-layer inter-leaved attention layers transformer $f_\theta$ for denoising feature embedding. Worth noting that, in each denoising step, only coarse level source point cloud $\hat{\M P}_t$ and its position encoding $\Theta(\hat{\M P}_t)$ (or target point cloud $\hat{\M Q}$ in 2D-3D registration task) have their values changed according to the warping operation, while other input parameters remain fixed. This is the key to our fast sampling speed.

\noindent\textbf{Attention Layer in $f_\theta$}: In the 3D registration task, following~\cite{li2022lepard}, the vectors $\M{q},\M{k},\M{v}$ in the self-attention, 
 are computed as:

\begin{equation}
\begin{aligned}
&\M q_i = \Theta(\M p_i)\mathbf{W}_qf^{\hat{\M p}_i}, \
\M k_j = \Theta(\M p_j)\mathbf{W}_kf^{\hat{\M p}_j}, \
\M v_j = \mathbf{W}_vf^{\hat{\M p}_j}, \\
&f^{\hat{\M p}_i} = f^{\hat{\M p}_i} + \rm{MLP}(\text{cat}[\M q_i,\Sigma_j\alpha_{ij}\M v_j]),
\end{aligned}\label{att_12}
\end{equation}
where $\mathbf{W}_q,\mathbf{W}_k,\mathbf{W}_v \in {\mathbb{R}}^{d{\times}d}$ are the attention weights, ${\alpha}_{ij} = softmax(\M q_{i}\M k_{j}^{\top}/{\sqrt{d}})$, and $\Theta(\cdot)$ is the relative rotationary position encoding~\cite{li2022lepard}. MLP$(\cdot)$ is a 3-layer fully connected network, and $cat[{\cdot},\cdot]$ is the concatenating operator. The cross attention layer is the standard form that $\M q$ and $\M k,\M v$ are computed by source and target point clouds, respectively. 

In the 2D-3D registration task, we take image inputs $\{\hat{\M X},\M F^{\hat{\M X}},\M F^{\hat{\M X}}_{dino}\}$ and point cloud input $\{\hat{\M Q},\M F^{\hat{\M Q}}\}$ to compute $\M q,\M k,\M v$ by utilizing standard attention layers~\cite{vaswani2017attention}. We also take Fourier embedding function~\cite{mildenhall2021nerf} to embed superpixels $\hat{\M X}$ and superpoints $\hat{\M Q}$ for positional encoding.

\noindent\textbf{Matching Function}: $\mathbf{matching\_logits}(\cdot,\cdot,\cdot,\cdot)$. \label{mlogits}
We compute matching ``logits'' between $\hat{\M P}$ and $\hat{\M Q}$ by features $\M F^{\hat{\M P}}$ (or $\M F^{\hat{\M X}}$) and $\M F^{\hat{\M Q}}$: $\tilde{\M E}(i,j) = \frac{1}{\sqrt{d}}\left<f^{\hat{\M p}_i}, f^{\hat{\M q}_j}\right>$. 

For the sake of clarity, we provide pseudo-code in Algorithm.\ref{denoising_algorithm} to describe the logic of our entire denoising module $g_{\theta}$ for the 3D Registration task. A similar definition of $g_{\theta}$ for the 2D-3D registration task can be found in the appendix.

\begin{algorithm}
\caption{Denoising Module $g_\theta$ for 3D Registration Task.}
\label{denoising_algorithm}
\begin{algorithmic}[1]
\Require Sampled matching matrix $\M E^t \in \mathbb{R}^{N\times M}$; Point clouds $\hat{\M P}, \hat{\M Q} \in \mathbb{R}^{3}$ and associated point features $\M F^{\hat{\M P}}, \M F^{\hat{\M Q}}$.
\Ensure Target matching matrix $\hat{\M E}_0$.
\Function{$g_{\theta}$}{$\M E^t, \hat{\M P}, \hat{\M Q}, \M F^{\hat{\M P}}, \M F^{\hat{\M Q}}$}
        \State $\tilde{\M E}_t \gets \M{f_{sinkhorn}}(\M E^t)$  
        \State $\M{\hat{R}}_t, \M{\hat{t}}_t \gets \M{soft\_procrustes}(\tilde{\M E}_t, \hat{\M P}, \hat{\M Q}), \ \hat{\M P}_t \gets \M{warping}(\hat{\M P}, \M{\hat{R}}_t, \M{\hat{t}}_t)$
        \State $\tilde{\M F}^{\hat{\M P}_t}, \tilde{\M F}^{\hat{\M Q}_t} \gets f_{\theta}(\hat{\M P}_t, \hat{\M Q}, \M F^{\hat{\M P}}, \M F^{\hat{\M Q}}, \Theta(\hat{\M P}_t), \Theta(\hat{\M Q}))$
        \State $\tilde{\M E}_0 \gets \M{matching\_logits}(\tilde{\M F}^{\hat{\M P}_t}, \tilde{\M F}^{\hat{\M Q}_t}, \Theta(\hat{\M P}_t), \Theta(\hat{\M Q}))$
        \State $\hat{\M E}_0 \gets \M{f_{sinkhorn}}(\tilde{\M E}_0)$
        \State \textbf{return} $\hat{\M E}_0$
\EndFunction
\end{algorithmic}
\end{algorithm}

\section{Experiments}\label{exp_section}

\subsection{3D Non-Rigid Registration Task}

\noindent\textbf{Datasets.}
4DMatch/4DLoMatch~\cite{li2022lepard} is an 3D non-rigid benchmark generated by the animation sequences from DeformingThings4D~\cite{li20214dcomplete}. We follow the dataset split provided in~\cite{li2022lepard}, which has a wide range of overlap ratio, that $45\%$-$92\%$ in 4DMatch and $15\%$-$45\%$ in 4DLoMatch.

\subsubsection{Implementation Details.}
Our framework utilize a KPConv~\cite{thomas2019kpconv} backbone to produce the superpoints $\hat{\M P}$ and $\hat{\M Q}$ and the asccociated features $\M F^{\hat{\M P}}$ and $\M F^{\hat{\M Q}}$. The dimension $d$ of superpoint features $\M F^{\hat{\M P}}$ and $\M F^{\hat{\M Q}}$ is set as $d = 432$.
Subsequently, we employ a repositioning transformer~\cite{li2022lepard} to provide a single-pass prediction of the matching matrix and the resulting transformation 
$[\M R, \M t]$, both of which are supervised by the matching loss $L_M$ and warping loss $L_W$ introduced in~\cite{li2022lepard}. We utilize a focal loss $L_{simple}$ (modified from Eqn.\ref{L_simple}) to guide the training of the denoising module $g_\theta$. The total loss function is defined as $L = L_M + L_W + L_{simple}$. 

We train the model for 30 epochs on the 4DMatch dataset with a batch size of 2. We adopt the training/validation/test split strategy from Predator~\cite{huang2021predator} and Lepard~\cite{li2022lepard}. At inference time, we conduct 20 iterations in the reverse sampling process while the total diffusion steps during training are set to 1000.

\noindent\textbf{Metrics.}
Following Lepard~\cite{li2022lepard}, we utilize two evaluation metrics to assess the quality of predicted matches. (1) Inlier Ratio (IR): The correct fraction in the correspondences prediction $\mathcal{K}_{pred}$. (2) Non-rigid Feature Matching Recall (NFMR): The fraction of ground truth correspondences $(u,v)\in \mathcal{K}_{gt}$ that can be successfully recovered by using the predicted correspondences $\mathcal{K}_{pred}$ as anchors. The NFMR metric provides a better characterization of the global rationality of overall body deformation, directly indicating whether the anchor $\mathcal{K}_{pred}$ effectively captures the body movements.

\begin{table}
\caption{Quantitative results on the 4DMatch and 4DLoMatch benchmarks. The best results are highlighted in bold, and the second-best results are underlined.}

\centering
	 \resizebox{0.65\textwidth}{!}{
\begin{tabular}{c|c|cc|cc}
\toprule
\midrule
\multirow{2}{*}{Category} & \multirow{2}{*}{Method} &  \multicolumn{2}{c}{4DMatch} & \multicolumn{2}{c}{4DLoMatch} \\
                  &                   &                     NFMR(\%)     &   IR(\%)       &      NFMR(\%)     &    IR (\%)     \\
\midrule

\multirow{3}{*}{Scene Flow} &           PointPWC~\cite{wu2019pointpwc}       & 21.60             &   20.0          &   10.0      &       7.20               \\
                  &    FLOT~\cite{puy2020flot}                          &  27.10    &     24.90          &  15.20         &    10.70         \\

\midrule                  
\multirow{4}{*}{Feature Matching} &       D3Feat~\cite{bai2020d3feat}                        &    55.50          &   54.70        &   27.40         &    21.50        \\
                  &        Predator~\cite{huang2021predator}                  &  56.40      &  60.40        &  32.10       &   27.50      \\
                 &       Lepard~\cite{li2022lepard}                       &  83.60       &    82.64       &     66.63      &   55.55       \\ 
                 &GeoTR~\cite{qin2022geometric} &83.20&82.20&65.40 &63.60\\                   
                 &RoITr~\cite{yu2023rotation}  &83.00 &\underline{84.40}&69.40 &\underline{67.60}\\

                &    Diff-Reg(Backbone)                &     \underline{85.47}   & 81.15     &     72.37   & 59.50      \\              &    Diff-Reg(steps=1)                               &    85.23    & 83.85     &   \underline{73.19}     & 65.26      \\   
                &    Diff-Reg(steps=20)                              &         \textbf{88.40}  &  \textbf{86.41}       &      \textbf{76.23}    &  \textbf{67.80}      \\
                           
\bottomrule        
\end{tabular}
}
\label{tab_4dmatch}

\end{table}
\noindent\textbf{Quantitative Results.}
We compare our method with two categories of state-of-the-art methods. The first category includes Scene Flow Methods such as PWC~\cite{wu2019pointpwc}, FLOT~\cite{puy2020flot}, and NSFP~\cite{li2021neural}. The second category encapsulates Feature Matching-Based Methods, namely D3Feat~\cite{bai2020d3feat}, Predator~\cite{huang2021predator}, Lepard~\cite{li2022lepard}, GeoTR~\cite{qin2022geometric}, and RoITr~\cite{yu2023rotation}. 

As illustrated in Table \ref{tab_4dmatch}, our method demonstrates significant improvements compared to the single-pass baselines. ``Diff-Reg(Backbone)'' refers to the single-pass prediction head (i.e., reposition transformer in Lepard~\cite{li2022lepard}), while ``Diff-Reg(steps=1)'' and ``Diff-Reg(steps=20)'' denotes our denoising module $g_{\theta}$ with one single step and 20 steps of reverse sampling. For both NFMR and IR metrics, ``Diff-Reg(steps=20)'' achieves the best performance. The improvement in NFMR of ``Diff-Reg(Backbone)'' compared to the baselines indicates that our diffused training samples in the matching matrix space enhance the feature backbone's representation, enabling the capture of crucial salient correspondences that are helpful for consistent global deformation. The significant enhancement of ``Diff-Reg(steps=20)'' over ``Diff-Reg(steps=1)'' demonstrates that the reverse denoising sampling process indeed searches for a better solution guided by the learned posterior distribution. 

To validate that the predicted correspondences indeed improve deformable registration, we conducted experiments using the state-of-the-art registration method GraphSCNet~\cite{qin2023deep}. As indicated in Table \ref{non_rigid_registration}, our predicted correspondences are beneficial for deformable registration, particularly in the more challenging 4DLoMatch benchmark.

\begin{table}
\caption{Non-rigid registration results of 4DMatch/4DLoMatch. Given predicted correspondences, we utilize the non-rigid registration method GraphSCNet~\cite{qin2023deep} to conduct the deformable registration. We retrain RoITr$^*$ using the authors' code. The modified 4DMatch-F and 4DLoMatch-F datasets~\cite{li2022non} exclude data involving near-rigid movements. The metrics (following~\cite{qin2023deep,li2022non}) are 3D End Point Error (EPE), 3D Accuracy Strict
(AccS) (<2.5cm or 5\%), 3D Accuracy Relaxed
(AccR)(<5cm or 5\%), and Outlier Ratio (OR) ( >30\%).}
\centering
\resizebox{0.8\textwidth}{!}{
\begin{tabular}{c|cccc|cccc}
\toprule
 \midrule
\multirow{2}{*}{Method}& \multicolumn{4}{c}{4DMatch-F} & \multicolumn{4}{c}{4DLoMatch-F} \\
 &   EPE$\downarrow$& AccS$\uparrow$ & AccR$\uparrow$ & OR$\downarrow$& EPE$\downarrow$& AccS$\uparrow$ & AccR$\uparrow$ & OR $\downarrow$     \\
 \midrule
 PointPWC~\cite{wu2019pointpwc} & 0.182& 6.25& 21.49& 52.07 &0.279 &1.69& 8.15& 55.70\\
 FLOT~\cite{puy2020flot} & 0.133& 7.66& 27.15 &40.49& 0.210 &2.73& 13.08& 42.51\\
 GeomFmaps [9]& 0.152 &12.34& 32.56 &37.90& 0.148& 1.85 &6.51 &64.63\\
 Synorim-pw [19] &0.099 &22.91& 49.86&26.01 &0.170 &10.55 &30.17& 31.12\\

Lepard~\cite{li2022lepard}$+$GraphSCNet~\cite{qin2023deep}&\underline{0.042}&70.10 &83.80& \underline{9.20}&  \underline{0.102}& \underline{40.00}& \underline{59.10}& \underline{17.50}  \\

 GeoTR~\cite{qin2022geometric}$+$GraphSCNet~\cite{qin2023deep}& 0.043 &\underline{72.10} &\underline{84.30}& 9.50& 0.119& 41.00& 58.40& 20.60   \\
RoITr$^*$~\cite{yu2023rotation} $+$GraphSCNet~\cite{qin2023deep}& 0.056 &59.60 &80.50& 12.50&0.118&32.30&56.70&20.50  \\
Diff-Reg$+$GraphSCNet~\cite{qin2023deep}& \textbf{0.041} &\textbf{73.20} &\textbf{85.80}& \textbf{8.30}& \textbf{0.095}& \textbf{43.80}& \textbf{62.90}& \textbf{15.50}   \\
 \bottomrule
\end{tabular}}
\label{non_rigid_registration}

\end{table}

\begin{figure*}
    \centering
    \begin{minipage}{\textwidth}
        \centering
        \includegraphics[width=0.15\textwidth,height=1.5cm]{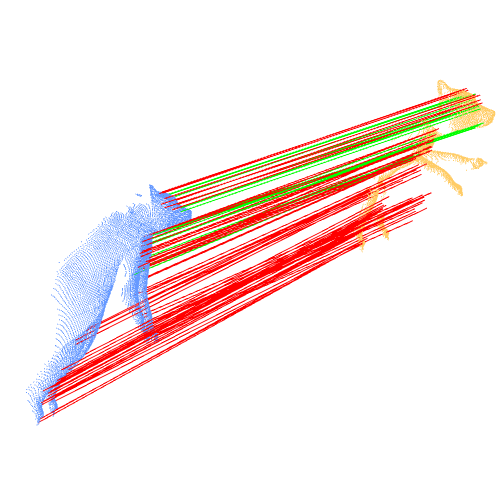}
        \hfill  
        \includegraphics[width=0.15\textwidth,height=1.5cm]{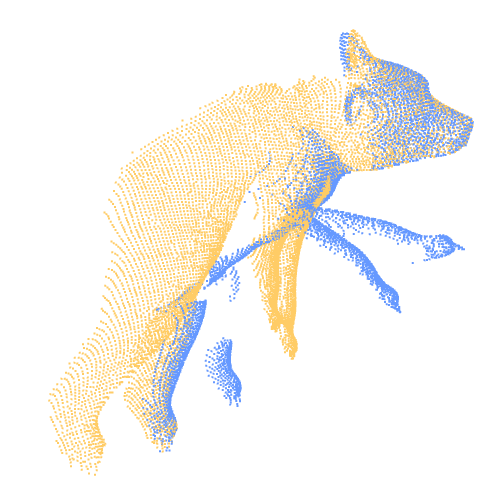}
        \hfill  
        \includegraphics[width=0.15\textwidth,height=1.5cm]{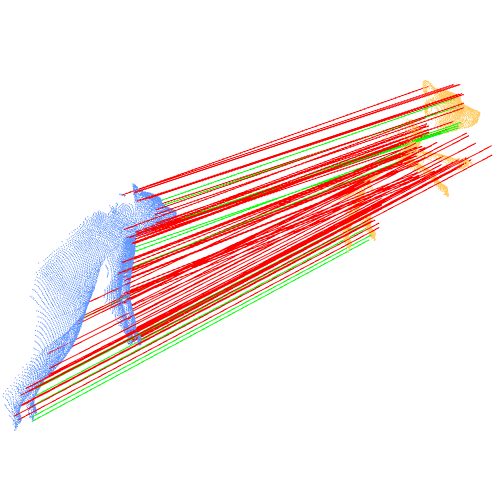}
        \hfill  
        \includegraphics[width=0.15\textwidth,height=1.5cm]{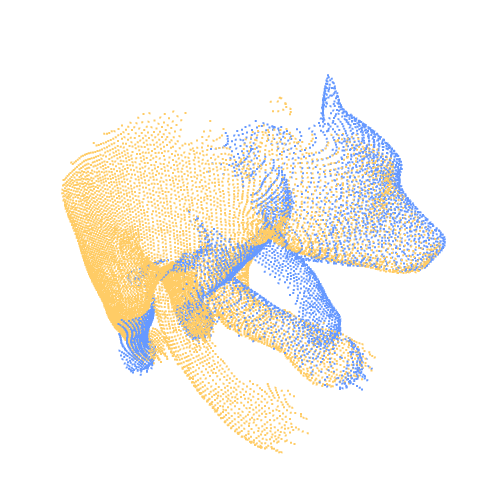} 
        \hfill  
        \includegraphics[width=0.15\textwidth,height=1.5cm]{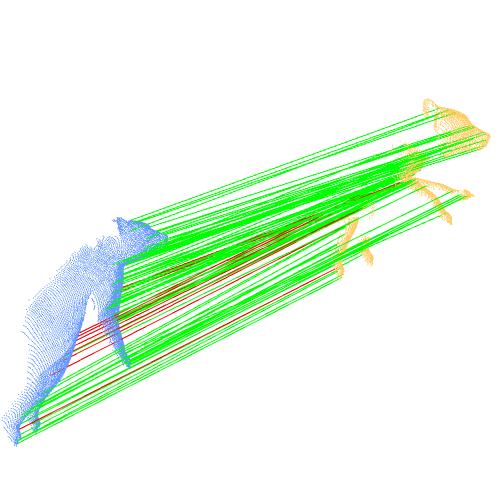} 
        \includegraphics[width=0.15\textwidth,height=1.5cm]
        {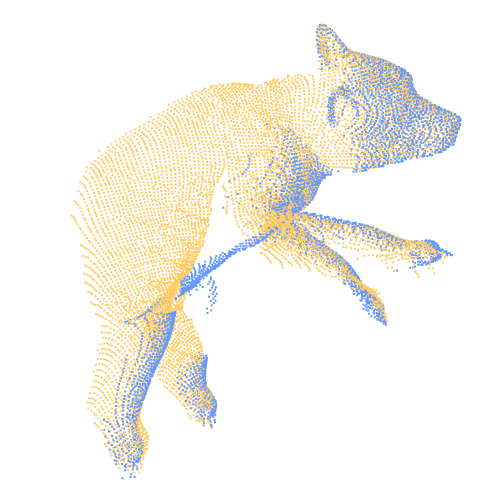} 
    \end{minipage}
  
    \begin{minipage}{\textwidth}

        \includegraphics[width=0.15\textwidth,height=1.5cm]{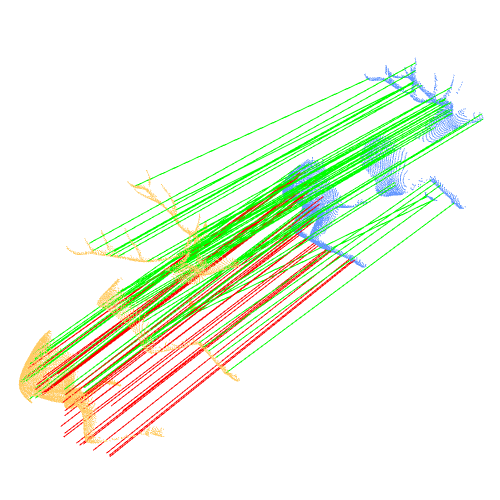}
        \hfill  
        \includegraphics[width=0.15\textwidth,height=1.5cm]{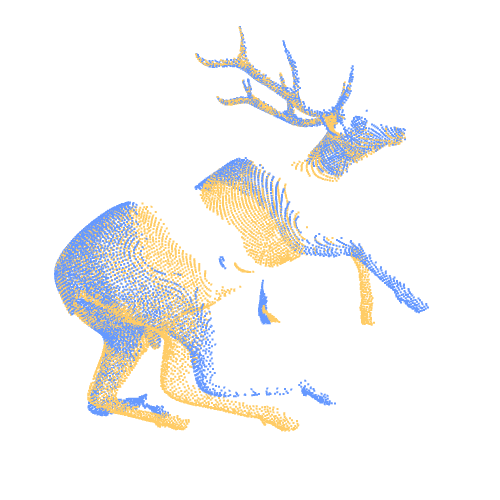}
        \hfill  
        \includegraphics[width=0.15\textwidth,height=1.5cm]{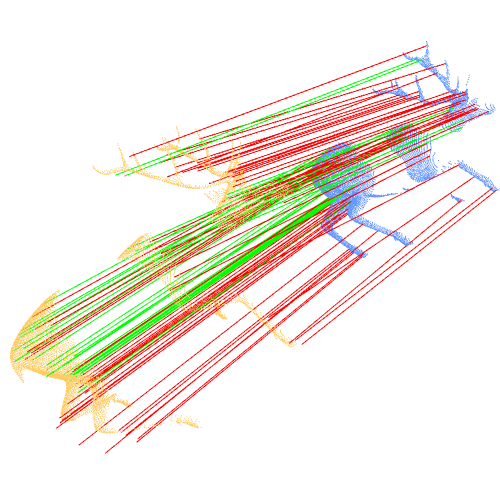}
        \hfill  
        \includegraphics[width=0.15\textwidth,height=1.5cm]{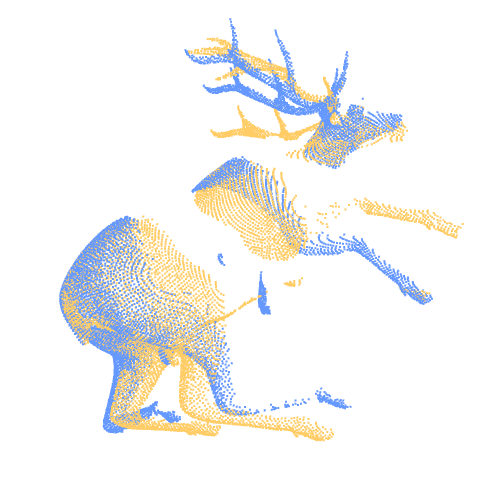} 
        \hfill  
        \includegraphics[width=0.15\textwidth,height=1.5cm]{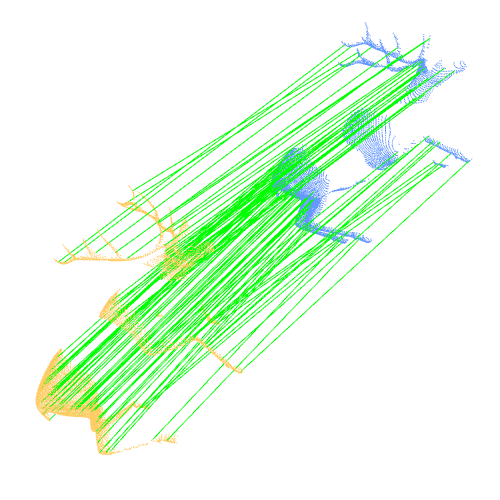} 
        \includegraphics[width=0.15\textwidth,height=1.5cm]
        {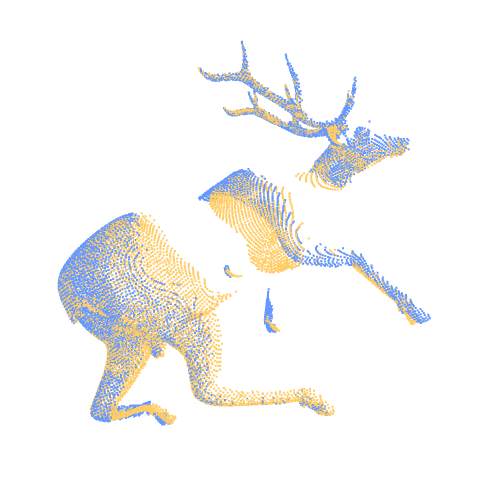} 
    \end{minipage}

    \begin{minipage}{\textwidth}
        \centering
        \includegraphics[width=0.15\textwidth,height=1.5cm]{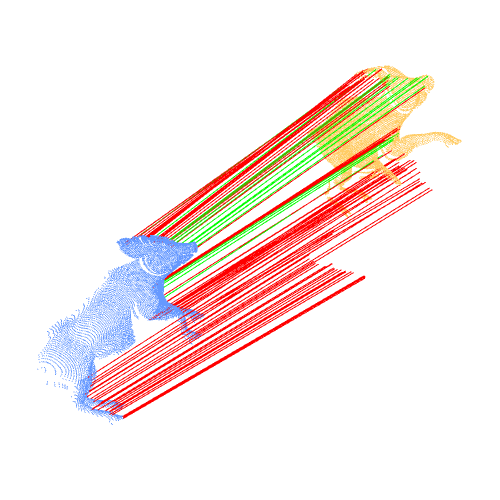}
        \hfill  
        \includegraphics[width=0.15\textwidth,height=1.5cm]{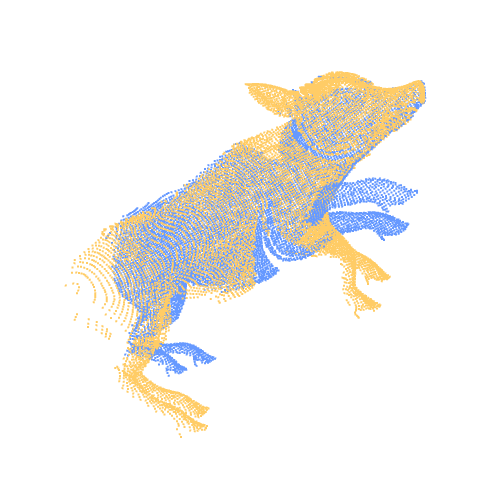}
        \hfill  
        \includegraphics[width=0.15\textwidth,height=1.5cm]{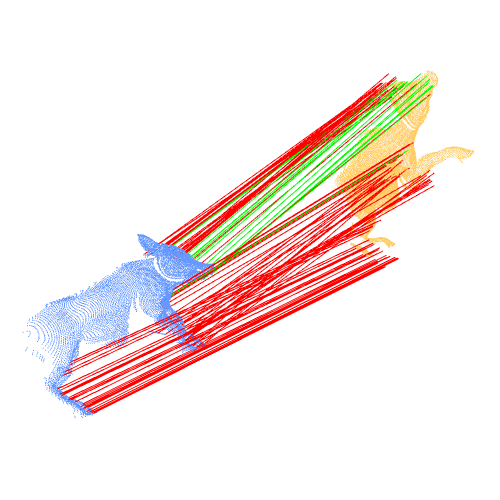}
        \hfill  
        \includegraphics[width=0.15\textwidth,height=1.5cm]{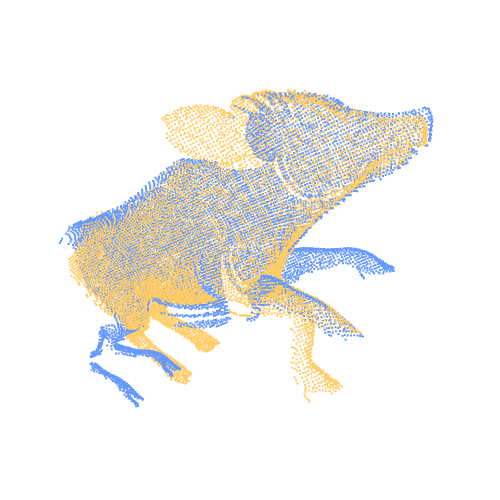} 
        \hfill  
        \includegraphics[width=0.15\textwidth,height=1.5cm]{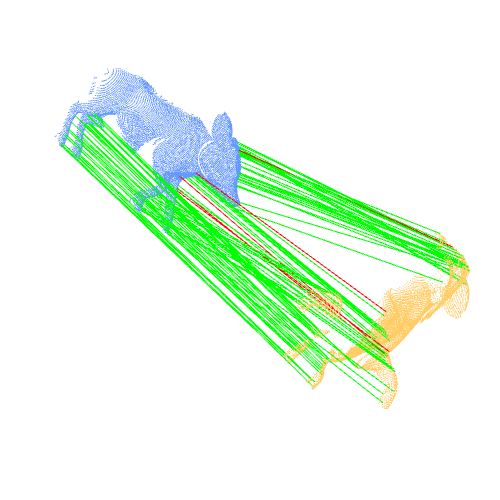} 
        \includegraphics[width=0.15\textwidth,height=1.5cm]{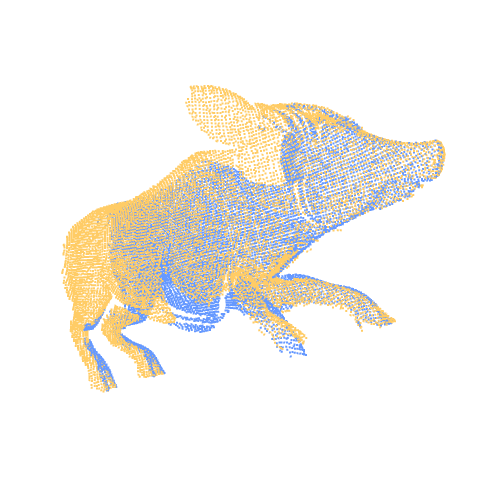}     
    \end{minipage}
    
    \begin{minipage}{\textwidth}
        \includegraphics[width=0.15\textwidth,height=1.5cm]{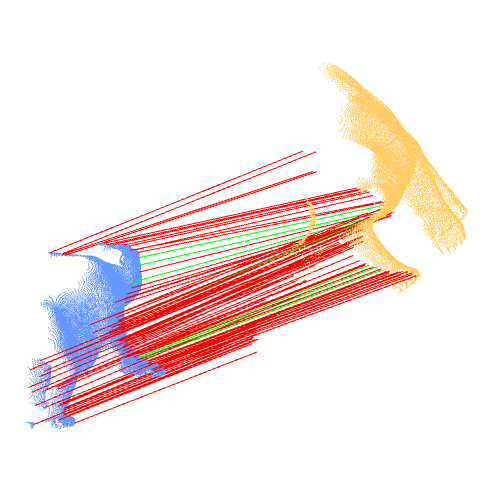}
        \hfill  
        \includegraphics[width=0.15\textwidth,height=1.5cm]{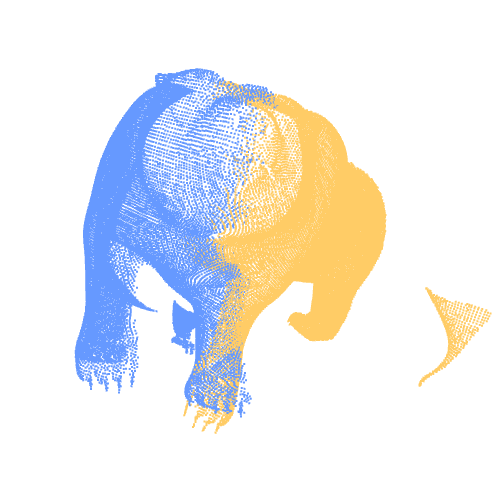}
        \hfill  
        \includegraphics[width=0.15\textwidth,height=1.5cm]{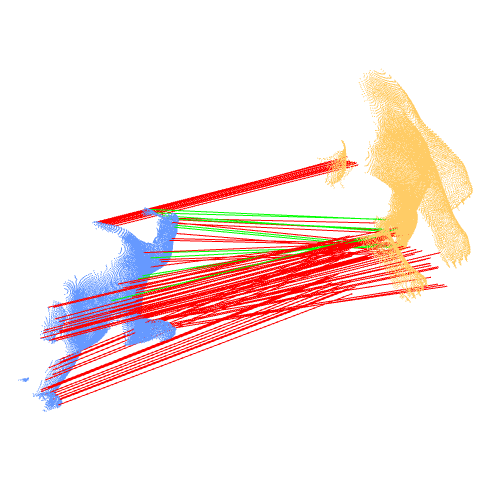}
        \hfill  
        \includegraphics[width=0.15\textwidth,height=1.5cm]{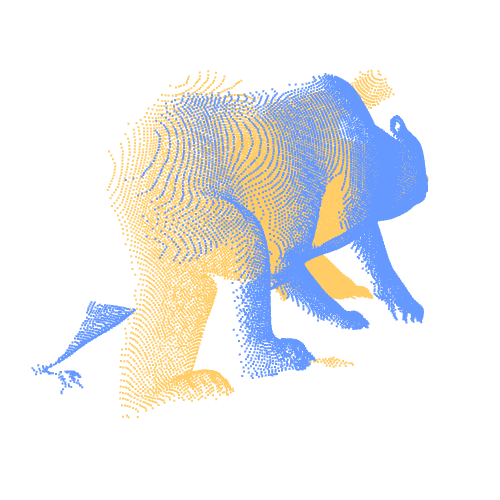} 
        \hfill  
        \includegraphics[width=0.15\textwidth,height=1.5cm]{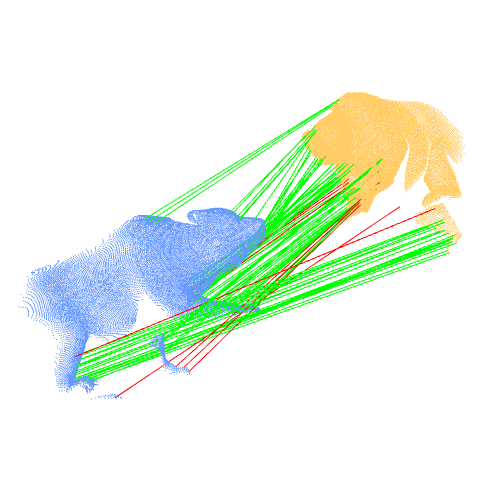} 
        \includegraphics[width=0.15\textwidth,height=1.5cm]{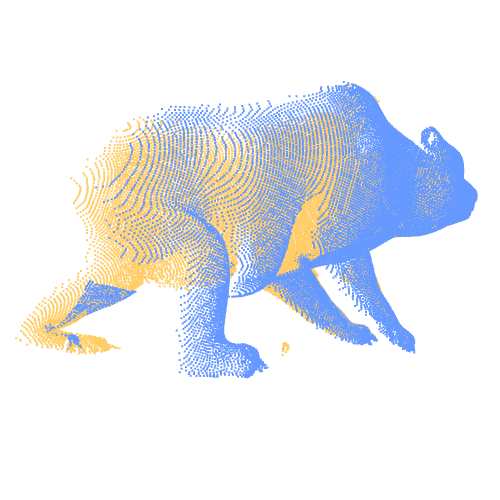}  
    \end{minipage}
    
    \begin{minipage}{\textwidth}
        \begin{subfigure}{0.3\textwidth}
            \includegraphics[height=1.5cm]{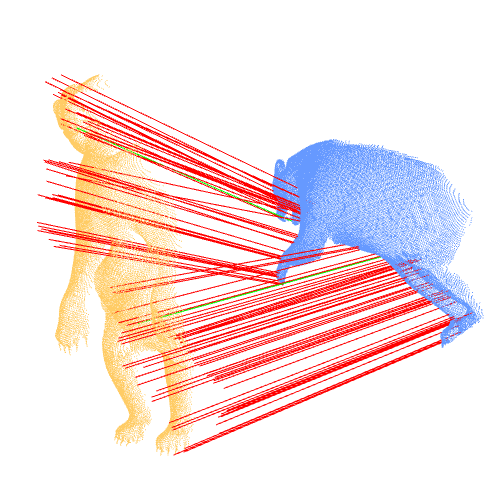}
            \hfill
            \includegraphics[height=1.5cm]{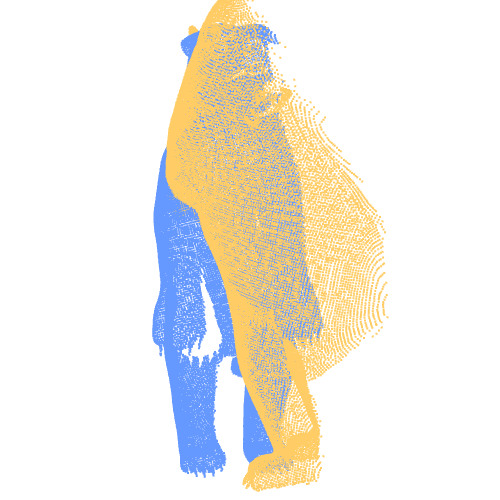}
        \caption{GeoTR\cite{qin2022geometric}}
        \end{subfigure}
        \hfill  
        \begin{subfigure}{0.3\textwidth}
            \includegraphics[height=1.5cm]{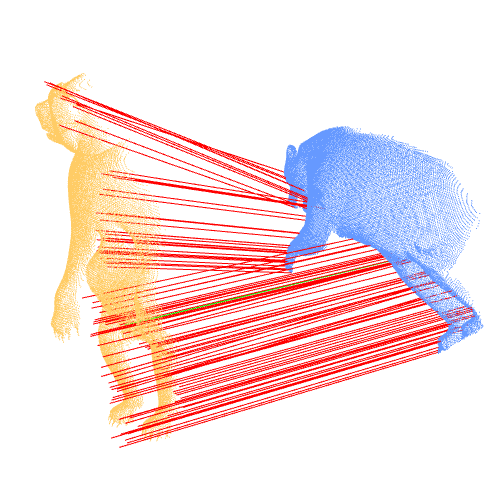}
            \hfill  
            \includegraphics[height=1.5cm]{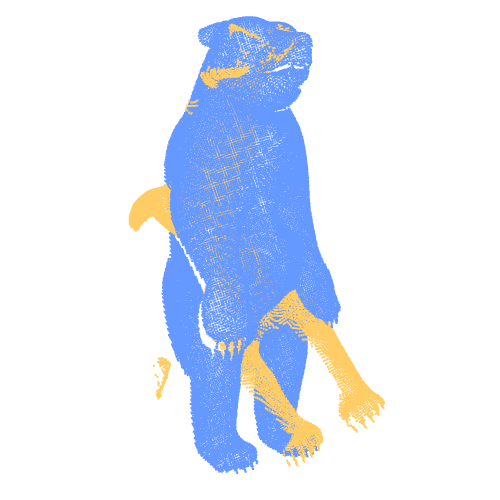} 
            \caption{RoITr\cite{yu2023rotation}}
        \end{subfigure}
        \hfill  
        \begin{subfigure}{0.3\textwidth}
            \includegraphics[height=1.5cm]{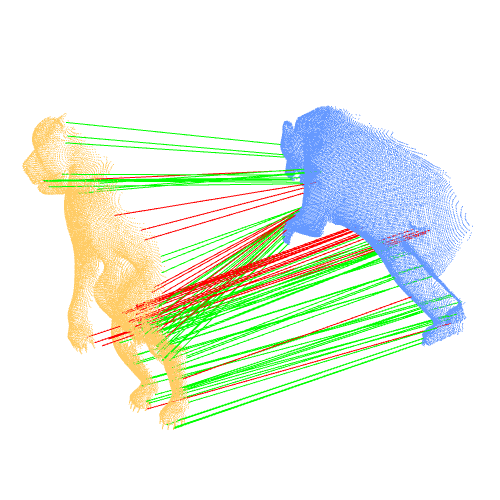} 
            \hfill
            \includegraphics[height=1.5cm]
            {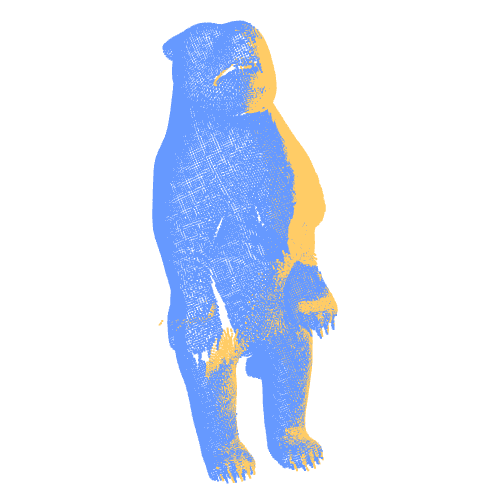}  
            \caption{Diff-Reg(steps=20)}
        \end{subfigure}            
        \label{4dmatchi_res_vis}
    \end{minipage}

    \caption{The qualitative results of non-rigid registration in the 4DMatch/4DLoMatch benchmark. The top two lines are from 4DMatch, while the bottom three are from 4DLoMatch. The \textcolor{blue}{blue} and \textcolor{yellow}{yellow} colors denote the source and target point cloud, respectively. The \textcolor{green}{green} and \textcolor{red}{red} lines indicate whether the predicted deformable flow from the source points is accepted by the threshold. The deformable registration is built by GraphSCNet~\cite{li20232d3d}. Zoom in for details.}
    \label{non_rigid_registration_vis}
\end{figure*}

\noindent\textbf{Qualitative Results.}
We provide a visualization to demonstrate our method's effectiveness in Fig.\ref{non_rigid_registration_vis}. For a fair comparison, we exploit the source point cloud's ``metric index'' (i.e., the test point set in the 4DMatch/4DLoMatch dataset) for all methods. Taking the predicted correspondences from RoITr~\cite{yu2023rotation}, GeoTr~\cite{yu2023rotation}, and ``Our (steps=20)'' as anchor correspondences, we calculate the deformation flow for the source test points by applying neighborhood k-nearest neighbors (KNN) interpolation based on the anchors. The deformable registration of the bear's two front paws in the first row and second/fourth column reveals that dealing with ambiguous matching patches of asymmetric objects can be highly challenging. However, our denoising process can handle this scenario perfectly. 

The deformable registration results from the top three rows (from the 4DMatch benchmark) indicate that the baseline methods struggle to compute reliable and consistent correspondences between scans with large deformations. The bottom two rows (from the 4DLoMatch benchmark) also demonstrate that low overlapping combined with deformation results in a disaster. These visualizations demonstrate that our denoising module in the matching matrix space provides a more effective approach for tackling deformable registration tasks.

\subsection{2D-3D Registration Task: RGB-D SCENES V2}
The 2D-3D registration task is non-trivial because the 2D image data is a perspective projection of the 3D scene, which creates the scale ambiguity problem~\cite{li20232d3d}. Since there is no good robust differentiable PnP~\cite{lepetit2009ep} solver that can be integrated into our denoising module, we deploy the SOTA depth estimation model DepthAnything~\cite{yang2024depth} to generate ``affine-invariant depth'' for the image. Subsequently, we utilize the camera's intrinsic matrix to project the estimated depth map onto a new non-metric point cloud paired with the corresponding real point cloud in the dataset, creating a challenge scale ambiguous registration problem. To alleviate the image data's scale ambiguity, we use the SOTA self-supposed pre-trained visual feature backbone DINOv2~\cite{oquab2023dinov2} to enhance the image features.

\noindent\textbf{Datasets.}
RGB-D Scenes V2~\cite{lai2014unsupervised} are generated from 14 indoor scenes comprising 11,427 RGB-D frames. Following~\cite{li20232d3d}, we split the 14 sequences into image-to-point-cloud pairs data, where scenes 0-8/11-14/9-10 are used for training/validation/testing. The resulting dataset contains 1,748 training pairs, 236 validation pairs, and 497 testing pairs of image-to-point-clouds.

\subsubsection{Implementation Details.}
For the single-pass backbone design, we follow 2D3D-MATR~\cite{li20232d3d}. Specifically, for images in data pair, we utilize a ResNet~\cite{he2016deep} with FPN~\cite{lin2017feature} to generate down-sampled superpixel and associated features. For the real point cloud in the data pair, we exploit KPConv~\cite{thomas2019kpconv} to extract the down-sampled superpoints with associated features. A transformer~\cite{vaswani2017attention} is deployed with inputs of superpixel and associated image features (including ResNet features and DINO features) from the image and superpoint and associated features from real point cloud to predict the cross-modality features. The denoising transformer in $g_{\theta}$ design has a similar definition. The new non-metric point cloud generated from the image depth map output by the DepthAnything model~\cite{yang2024depth} is used as inputs of the weighted SVD function in $g_{\theta}$ to compute $\M{R},\M{t}$. 

We utilize the coarse level circle loss~\cite{li20232d3d} and fine level matching loss~\cite{li20232d3d} for the singe-pass backbone~\cite{li20232d3d}, while a focal loss for our denoising module $g_{\theta}$. More details about network design are in the appendix. We train our model about 30 epochs with batch size 1.

\noindent\textbf{Metrics.}
We evaluate our method using the Registration Recall (RR) metric: the ratio of image-to-point-cloud pairs' RMSE is under 10cm.

\begin{table}[H]
\centering
\caption{Evaluation results on RGB-D Scenes V2~\cite{li20232d3d}. The best results are highlighted in bold, and the second-best results are underlined.}
\resizebox{0.7\textwidth}{!}{
\begin{tabular}{c|ccccc}
\toprule
\midrule
Method & Scene-11 &Scene-12 &Scene-13 &Scene-14 &Mean \\
\midrule
Mean depth (m) &1.74& 1.66 &1.18 &1.39 &1.49\\
\midrule
\multicolumn{5}{c}{Registration Recall(\%)$\uparrow$}\\
\midrule
FCGF-2D3D~\cite{choy2019fully} &26.4& 41.2& 37.1 &16.8 &30.4\\
P2-Net~\cite{Wang2021P2NetJD} &40.3& 40.2& 41.2& 31.9& 38.4\\
Predator-2D3D~\cite{huang2021predator} &44.4 &41.2& 21.6 &13.7& 30.2\\
2D3D-MATR~\cite{li20232d3d}& 63.9 &53.9 &58.8& 49.1& 56.4\\
FreeReg+Kabsch~\cite{Wang2023FreeRegIC}& 38.7& 51.6& 30.7& 15.5& 34.1 \\
FreeReg+PnP~\cite{Wang2023FreeRegIC}& 74.2& 72.5& 54.5& 27.9& 57.3\\
\midrule
Diff-Reg(dino) &87.5&86.3&63.9&60.6&74.6\\
Diff-Reg(dino/backbone) &79.2&86.3&75.3&\textbf{71.2}&78.0\\
Diff-Reg(dino/steps=1) &\underline{94.4}&\underline{98.0}&\underline{85.6}&\underline{63.7}&\underline{85.4}\\
Diff-Reg(dino/steps=10) &\textbf{98.6}&\textbf{99.0}&\textbf{86.6}&\underline{63.7}&\textbf{87.0}\\
\midrule
Diff-Reg(dino/backbone$_{epnp}$) &95.8&96.1&88.7&69.0&87.4\\
\bottomrule
\end{tabular}
}
\label{tab:non_rigid_registration}
\end{table}

\noindent\textbf{Quantitative Results.}
We introduce a single-pass baseline ``Diff-Reg(dino)'', in which we integrate the visual foundation model DINOv2~\cite{oquab2023dinov2} into the single-pass model 2D3D-MATR's~\cite{li20232d3d} ResNet. ``Diff-Reg(dino/backbone)'' represents the single-pass prediction head derived from ``Diff-Reg(dino)'' after joint training with our denoising module $g_{\theta}$. ``Diff-Reg(steps=1)'' and ``Diff-Reg(steps=10)'' refer to our diffusion matching model with one step and ten steps of reverse denoising sampling.

As demonstrated in Table \ref{tab:non_rigid_registration}, the results for ``Diff-Reg(dino/backbone)'' indicate that the diffused training samples in the matrix space serve as data augmentation to enhance the representation of the ResNet feature backbone and single-pass prediction head. Additionally, the outcome for ``Diff-Reg(dino/steps=10)'' reveals that our denoising module $g_{\theta}$ effectively tackles the scale ambiguous issue in the 2D-3D registration. Furthermore, the result for ``Diff-Reg(dino/steps=1)'' reveals that the diffused training samples within the matrix space indeed enhance the single-pass prediction head in ``Diff-Reg(dino).''

We also carried out an additional experiment to show that the performance improvement of ``Diff-Reg(dino/backbone)'' compared to ``Diff-Reg(dino)'' is attributed to our diffusion matching model. In this experiment, we did not use the depth map (from DepthAnything model\cite{yang2024depth}), opting instead to employ a differentiable weighted EPnP~\cite{lepetit2009ep} solver in the denoising module $g_{\theta}$ to replace the weighted SVD layer. We take only superpixels and superpoints as inputs of the EPnP solver, with correspondence weights computed by associated superpixel and superpoint features. This setting denoted by ``Diff-Reg(dino/backbone$_{epnp}$)'' (in Table~\ref{tab:non_rigid_registration}) achieved a high recall rate of 87.4\%, indicating that our diffused samples in matching matrix space indeed enhance the feature backbone.

\begin{figure*}
    \centering
    \subfloat[Diff-Reg(dino)]{\begin{tabular}[b]{c}
    \includegraphics[width=0.31\linewidth, height=1.5cm]{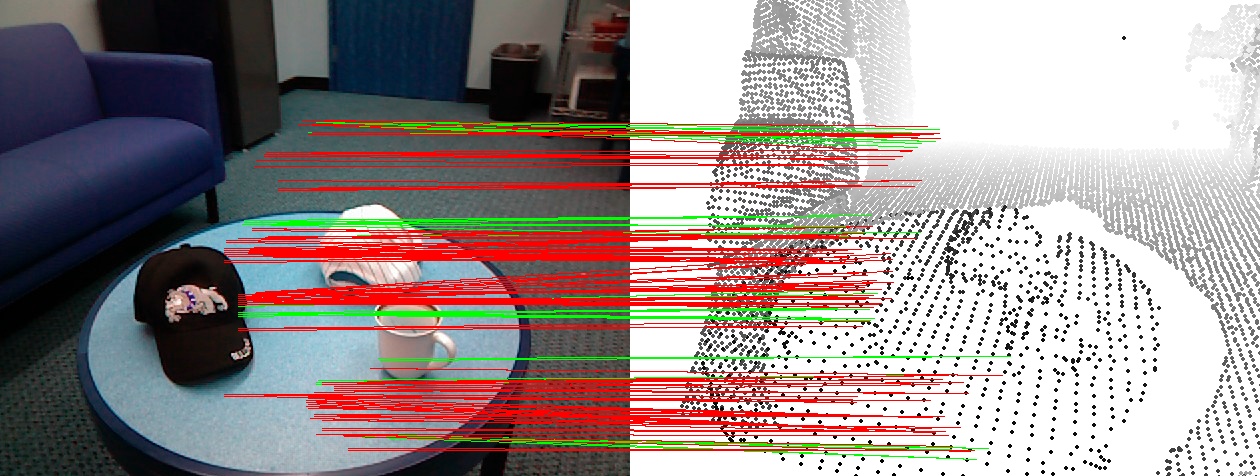} \\ \includegraphics[width=0.31\linewidth, height=1.5cm]{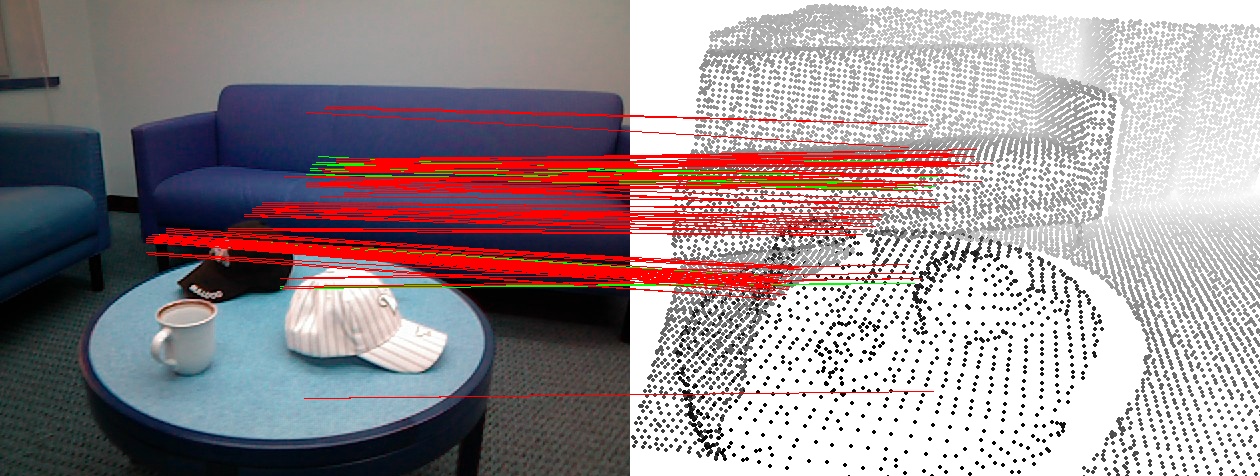}\\
    \includegraphics[width=0.31\linewidth, height=1.5cm]{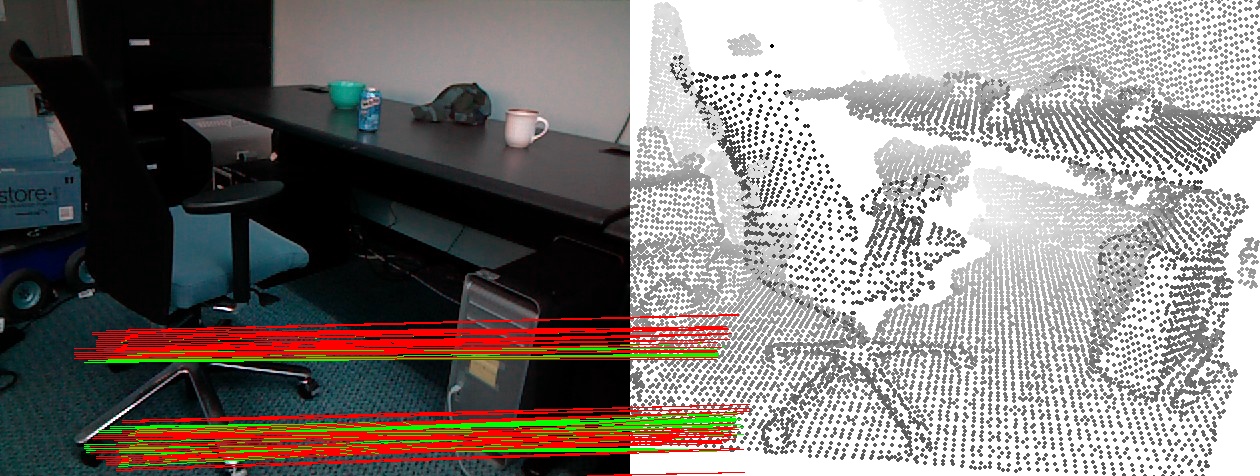}\\
    \includegraphics[width=0.31\linewidth, height=1.5cm]{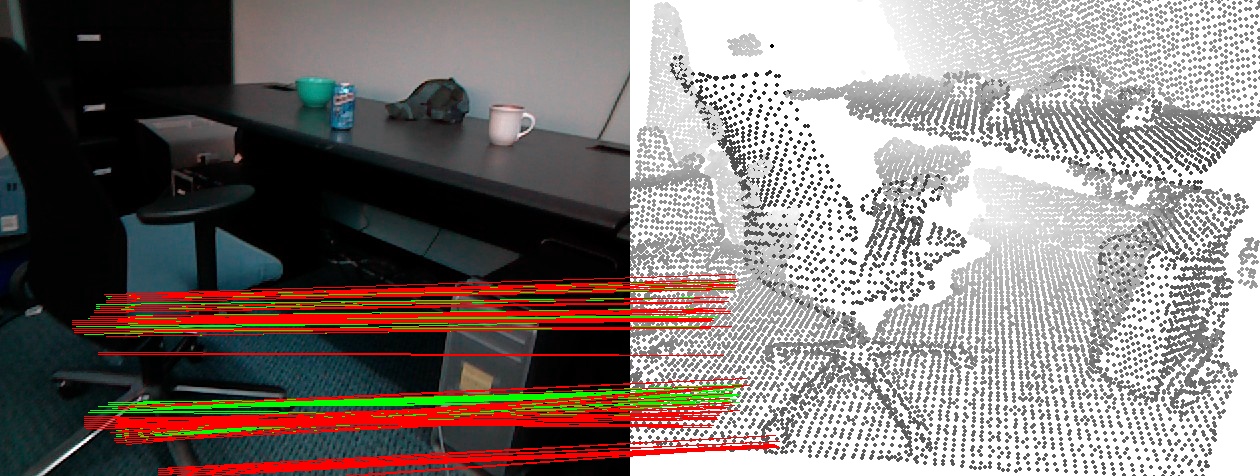}
    \end{tabular}} \quad
    \subfloat[Diff-Reg(dino/steps=10)]{\begin{tabular}[b]{c}
    \includegraphics[width=0.31\linewidth, height=1.5cm]{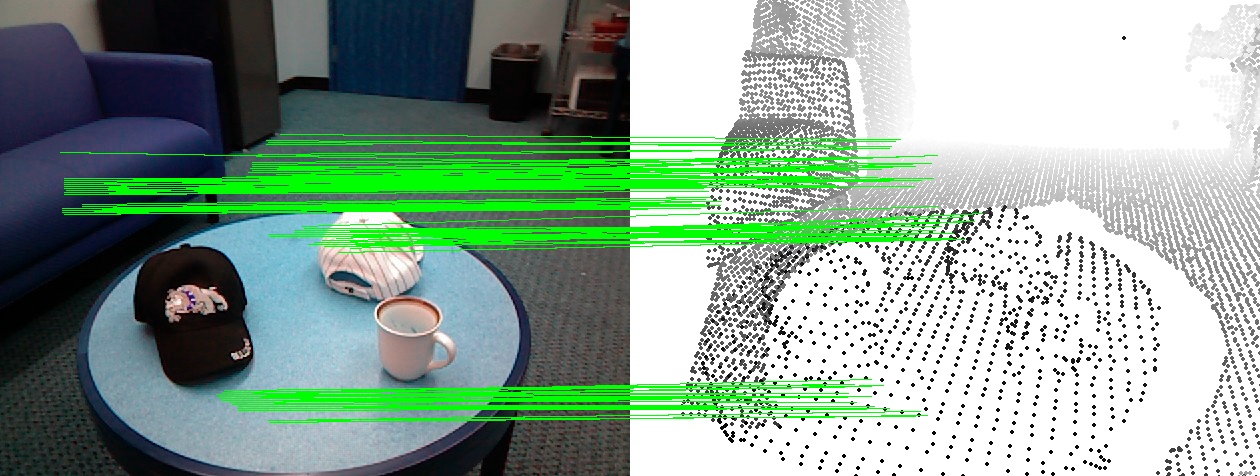}
    \includegraphics[width=0.31\linewidth, height=1.5cm]{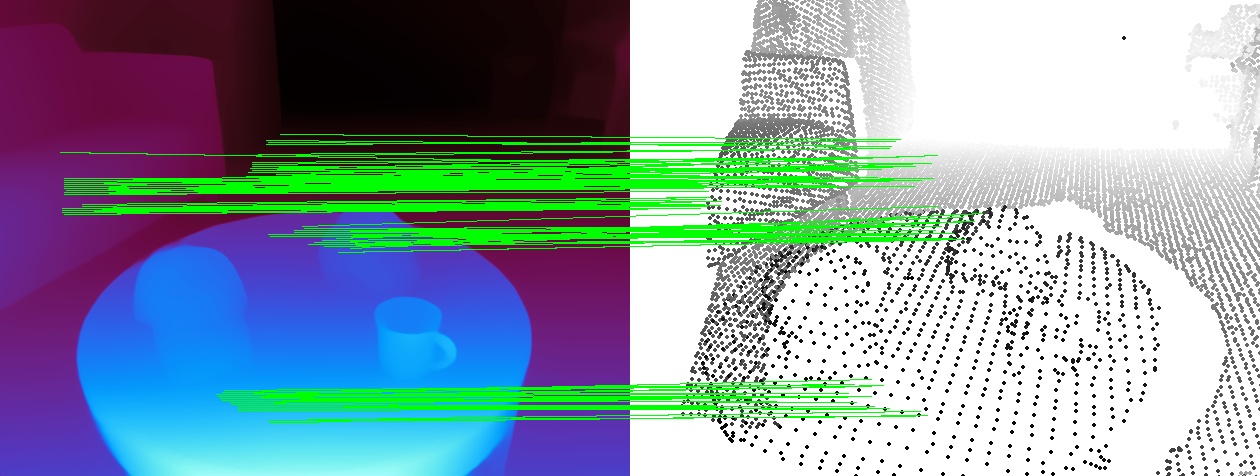} \\ \includegraphics[width=0.31\linewidth, height=1.5cm]{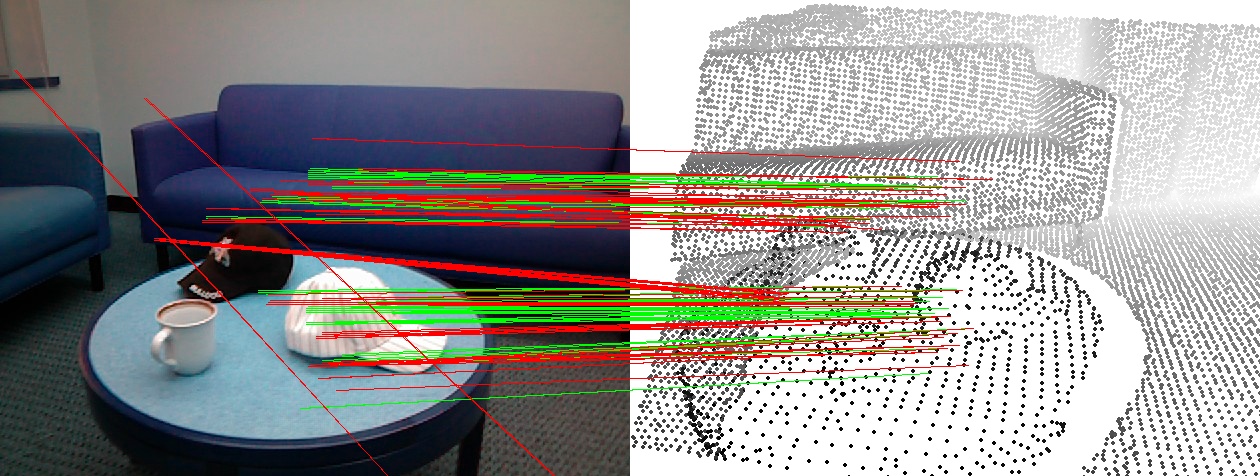}
    \includegraphics[width=0.31\linewidth, height=1.5cm]{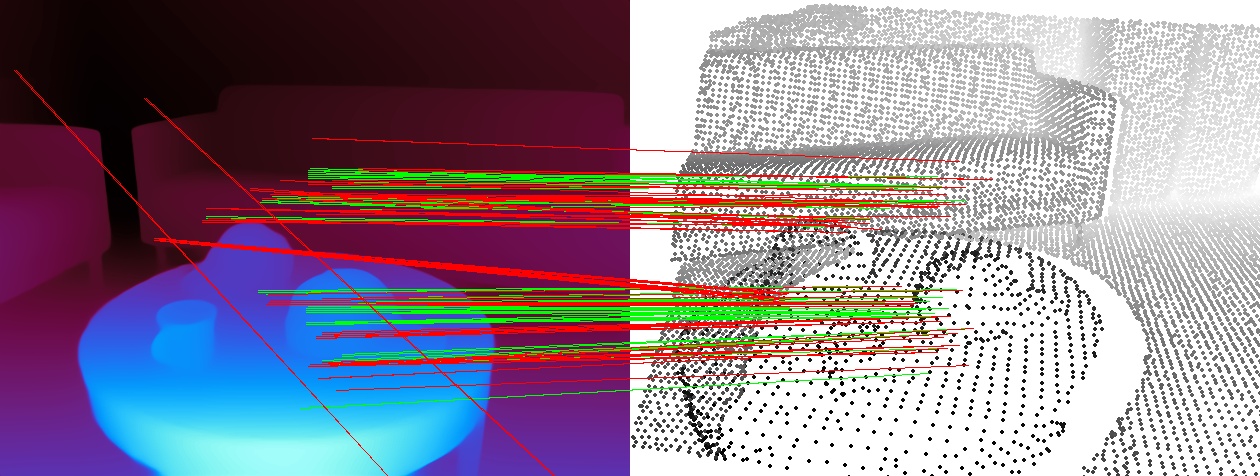}\\
    \includegraphics[width=0.31\linewidth, height=1.5cm]{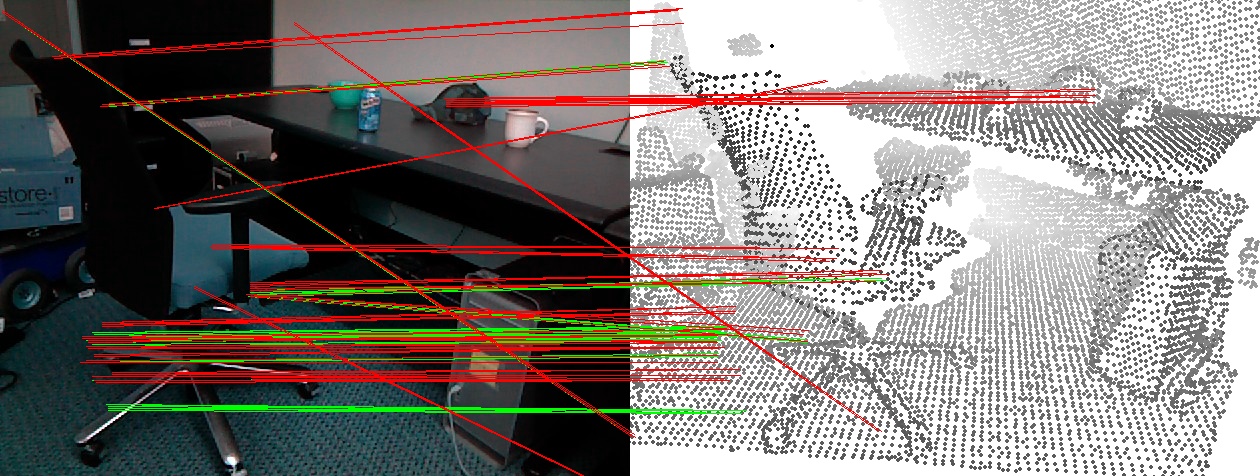}
    \includegraphics[width=0.31\linewidth, height=1.5cm]{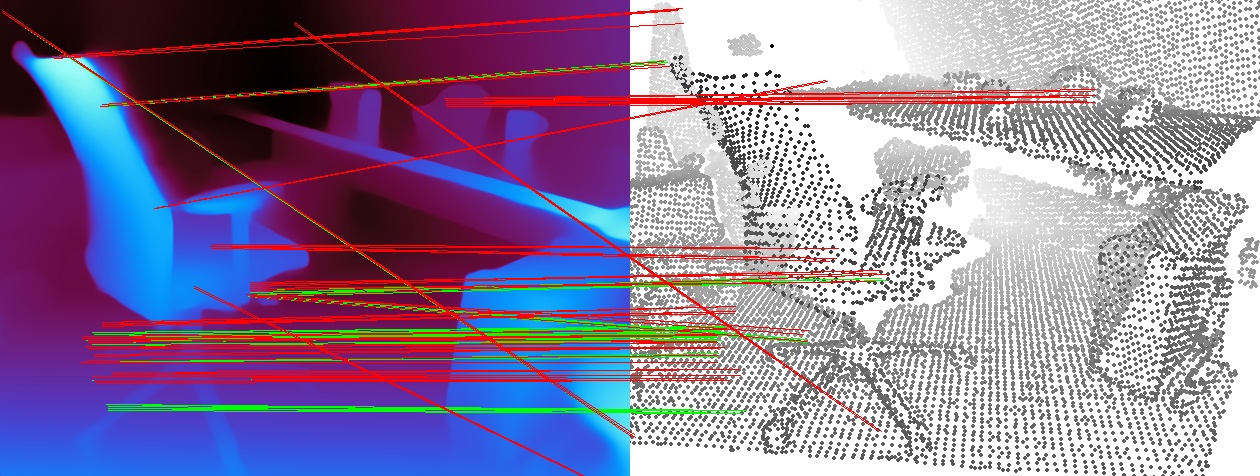}\\
    \includegraphics[width=0.31\linewidth, height=1.5cm]{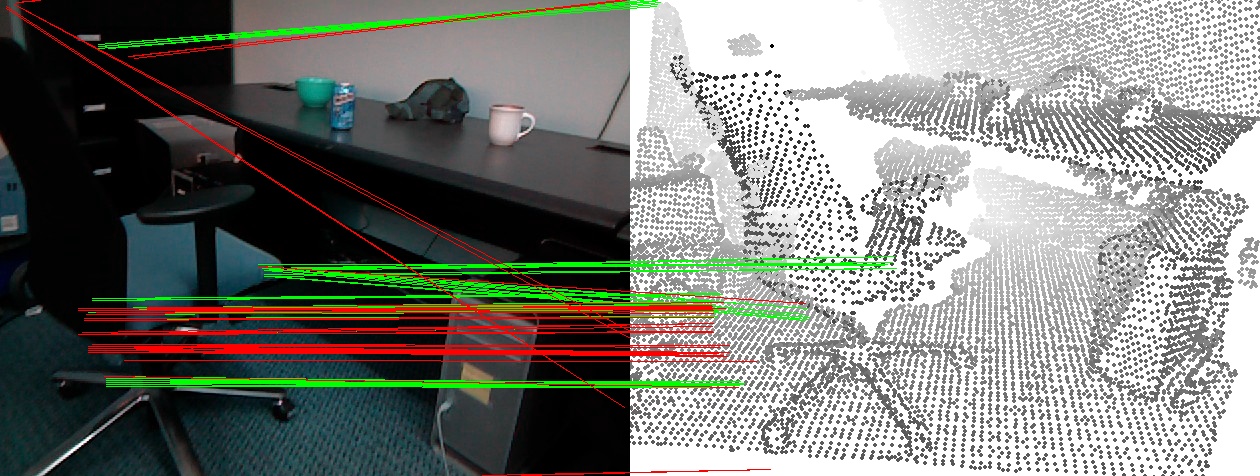}
    \includegraphics[width=0.31\linewidth, height=1.5cm]{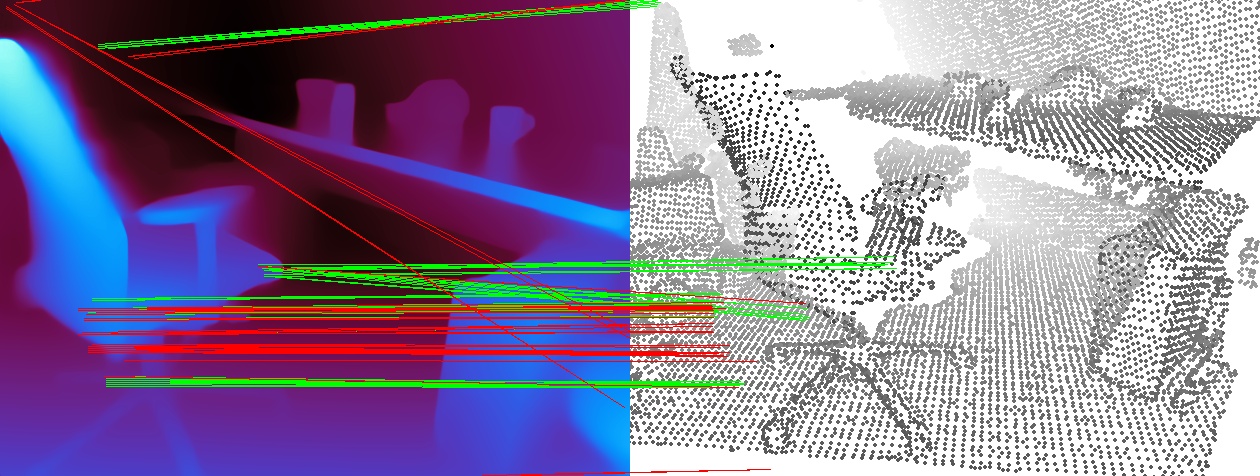}
    \end{tabular}}
    \caption{The qualitative results of top-200 predicted correspondences on the RGB-D Scenes V2 benchmark~\cite{lai2014unsupervised}. The \textcolor{green}{green}/\textcolor{red}{red} color indicates whether the matching score is accepted based on a threshold value. Zoom in for details.}
	\label{2d3d_non_rigid_registration_vis}
\end{figure*}

\noindent\textbf{Qualitative Results.}
The prediction examples of "Diff-Reg(dino/steps=10)" in Fig. \ref{2d3d_non_rigid_registration_vis} reveal that our diffusion matching model excels at capturing salient correspondences crucial for combinatorial consistency, regardless of their distance from the camera. On the other hand, the matches generated by ``Diff-Reg(dino)'' tend to be more focused at specific distances. For instance, in the third row, the correspondences produced by ``Diff-Reg(dino)'' are located very close to the camera, while the correspondences from ``Diff-Reg(dino/steps=10)'' encompass objects such as hats on the black table that are situated at a greater distance. In the first row, ``Diff-Reg(dino)'' fails to capture the correspondences on the sofa, and in the second row, the correspondences of the white hat on the table are lost. An extreme case in the fourth row demonstrates that ``Diff-Reg(dino)'' misses the farthest correspondence on the column bookshelf or wall.

\section{Conclusion}
This paper presents a novel diffusion module that leverages a diffusion matching model in the doubly stochastic matrix space to learn a posterior distribution for guiding the reverse denoising sampling process within the matrix space. Moreover, we have integrated a lightweight design into the denoising module to decrease the time cost associated with iterative reverse sampling. Experimental results on both 3D registration and 2D-3D registration tasks confirm the effectiveness and efficiency of our proposed denoising module.
\bigskip

\clearpage

\bibliographystyle{splncs04}
\bibliography{root}
\end{document}